\documentclass[11pt]{article}

\usepackage{ACL2023}

\usepackage{times}
\usepackage{latexsym}
\usepackage{graphicx}
\usepackage{todonotes}
\usepackage{tcolorbox}
\usepackage{multirow}
\usepackage{colortbl}
\usepackage{amsmath}

\usepackage[T1]{fontenc}

\usepackage[utf8]{inputenc}

\usepackage{microtype}
\usepackage{multirow}
\usepackage{inconsolata}
\usepackage{caption}
\usepackage{subcaption}
 \usepackage{tabularray}

\usepackage{scalerel,xparse}

\NewDocumentCommand\emojititle{}{
    \scalerel*{
        \includegraphics{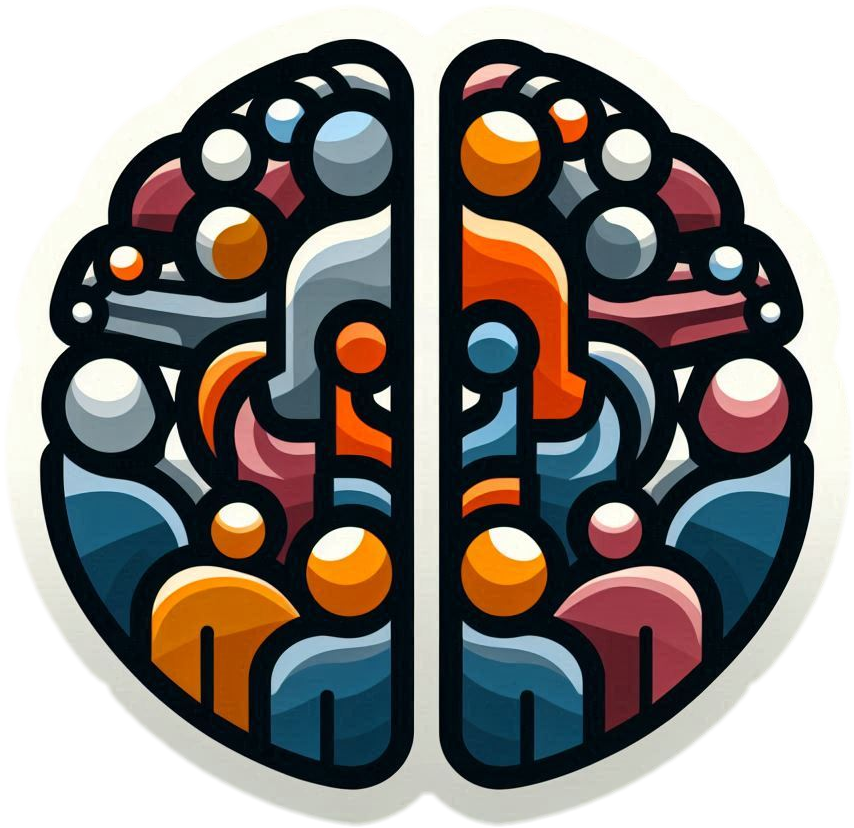}
    }{\rule[-3pt]{1pt}{0.65cm}}
}

\NewDocumentCommand\emojifrance{}{
    \scalerel*{
        \includegraphics{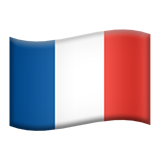}
    }{\rule[-3pt]{1pt}{0.47cm}}
}

\NewDocumentCommand\emojibulgaria{}{
    \scalerel*{
        \includegraphics{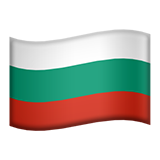}
    }{\rule[-3pt]{1pt}{0.47cm}}
}

\NewDocumentCommand\emojiitaly{}{
    \scalerel*{
        \includegraphics{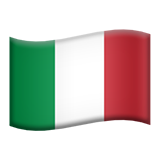}
    }{\rule[-3pt]{1pt}{0.47cm}}
}

\NewDocumentCommand\emojipoland{}{
    \scalerel*{
        \includegraphics{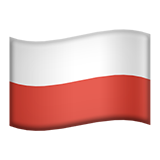}
    }{\rule[-3pt]{1pt}{0.47cm}}
}

\NewDocumentCommand\emojislovakia{}{
    \scalerel*{
        \includegraphics{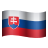}
    }{\rule[-3pt]{1pt}{0.47cm}}
}

\NewDocumentCommand\emojirobot{}{
    \scalerel*{
        \includegraphics{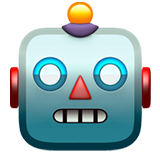}
    }{\rule[-3pt]{1pt}{0.47cm}}
}

\NewDocumentCommand\emojiuk{}{
    \scalerel*{
        \includegraphics{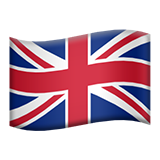}
    }{\rule[-3pt]{1pt}{0.47cm}}
}

\NewDocumentCommand\emojiczechia{}{
    \scalerel*{
        \includegraphics{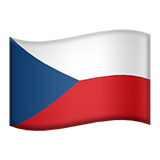}
    }{\rule[-3pt]{1pt}{0.47cm}}
}

\NewDocumentCommand\emojigermany{}{
    \scalerel*{
        \includegraphics{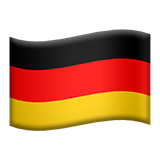}
    }{\rule[-3pt]{1pt}{0.47cm}}
}

\NewDocumentCommand\emojihungary{}{
    \scalerel*{
        \includegraphics{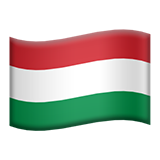}
    }{\rule[-3pt]{1pt}{0.47cm}}
}

\NewDocumentCommand\emojilithuania{}{
    \scalerel*{
        \includegraphics{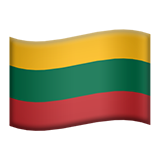}
    }{\rule[-3pt]{1pt}{0.47cm}}
}

\NewDocumentCommand\emojispain{}{
    \scalerel*{
        \includegraphics{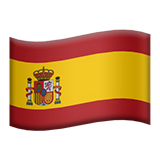}
    }{\rule[-3pt]{1pt}{0.47cm}}
}

\NewDocumentCommand\emojigamedie{}{
    \scalerel*{
        \includegraphics{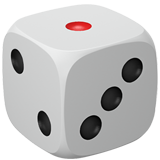}
    }{\rule[-3pt]{1pt}{0.47cm}}
}

\NewDocumentCommand\emojicheckmark{}{
    \scalerel*{
        \includegraphics{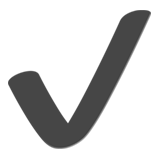}
    }{\rule[-3pt]{1pt}{0.47cm}}
}

\NewDocumentCommand\emojicrossmark{}{
    \scalerel*{
        \includegraphics{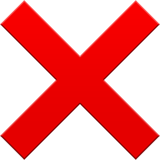}
    }{\rule[-3pt]{1pt}{0.47cm}}
}

\title{\emojititle Persona Inconstancy in Multi-Agent LLM Collaboration: \\Conformity, Confabulation, and Impersonation}

\author{Razan Baltaji\textsuperscript{1}, Babak Hemmatian\textsuperscript{2}, Lav R.\ Varshney\textsuperscript{1,2}\\
        \textsuperscript{1}Department of Electrical and Computer Engineering\\ \textsuperscript{2}Beckman Institute for Advanced Science and Technology \\ University of Illinois Urbana-Champaign\\
        \texttt{\{baltaji, babak2, varshney\}@illinois.edu}}
        
\begin{document}
\maketitle

\begin{abstract}

Multi-agent AI systems can be used for simulating collective decision-making in scientific and practical applications. They can also be used to introduce a diverse group discussion step in chatbot pipelines, enhancing the cultural sensitivity of the chatbot's responses. These applications, however, are predicated on the ability of AI agents to reliably adopt assigned personas and mimic human interactions. To see whether LLM agents satisfy these requirements, we examine AI agent ensembles engaged in cross-national collaboration and debate by analyzing their private responses and chat transcripts. Our findings suggest that multi-agent discussions can support collective AI decisions that more often reflect diverse perspectives, yet this effect is tempered by the agents' susceptibility to conformity due to perceived peer pressure and occasional challenges in maintaining consistent personas and opinions. Instructions that encourage debate in support of one's opinions rather than collaboration increase the rate of inconstancy. Without addressing the factors we identify, the full potential of multi-agent frameworks for producing more culturally diverse AI outputs or more realistic simulations of group decision-making may remain untapped. 


\end{abstract}

\noindent \textcolor{red}{ Warning: Contains potentially unsafe LLM responses.}

\section{Introduction}
\begin{figure}[!t]
\centering
\includegraphics[width=\columnwidth]{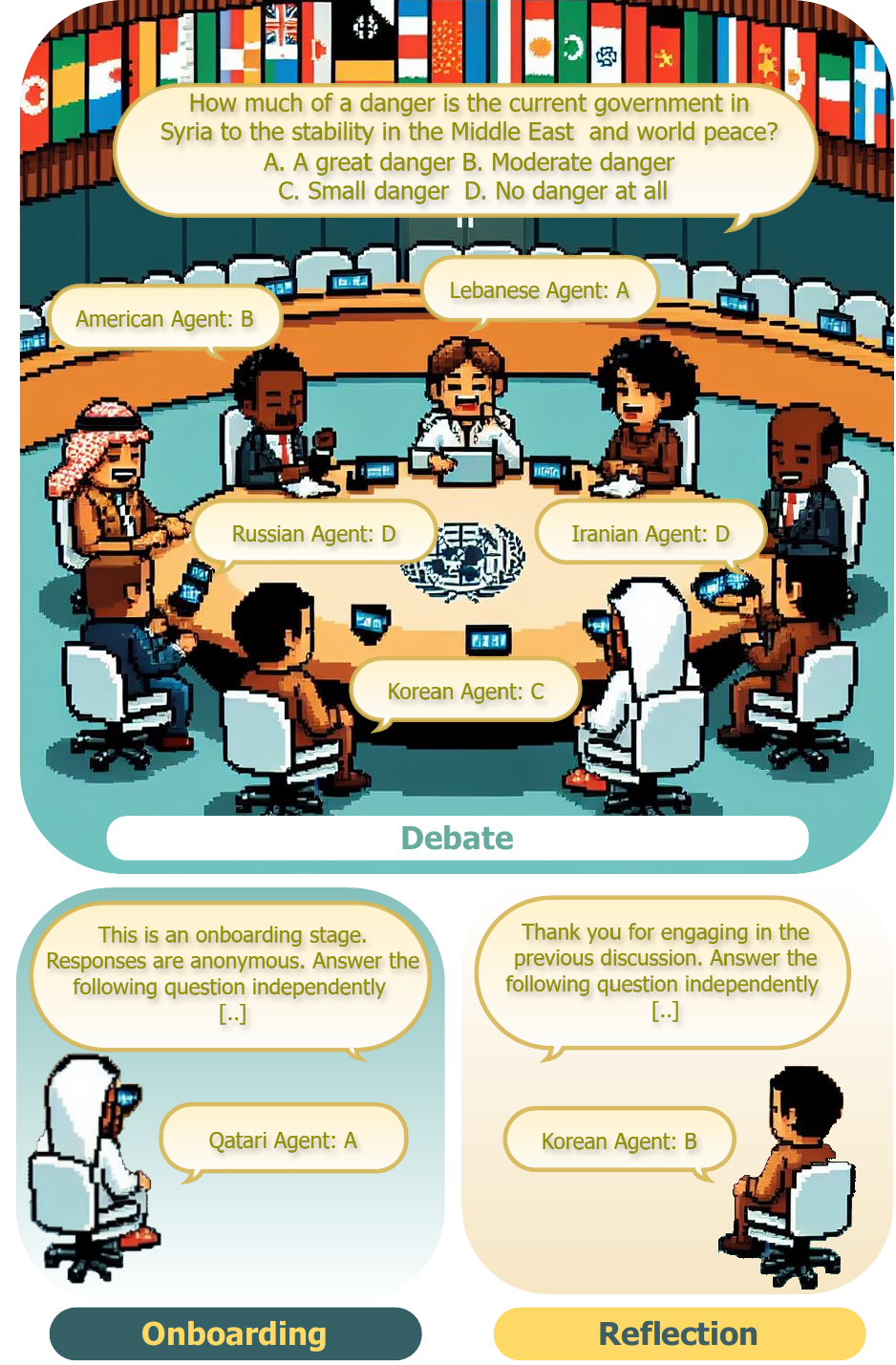}
\label{schematic}
\caption{An illustration of our experimental setup for a debate: a) \textit{Onboarding} stage where agents are asked to report their opinions independently, b) \textit{Debate} stage where agents participate in a debate moderated by a chat manager, c) \textit{Reflection} stage where agents are asked to report their opinions independently based on the previous discussion. A similar setup is used for collaboration.}
\end{figure}

The functioning of institutions throughout a democratic society depends on fruitful group discussions and the consequent collective decision-making. These interactions determine everything from the policies a school may enact, to the laws that determine national and even international conduct. Therefore, it is unsurprising that much research in the behavioral sciences over the past century has focused on understanding the dynamics of group interactions, and what these dynamics mean for discursive or behavioral outcomes (for a contemporary overview, see \citet{brown2019group}). Some of the prominent questions asked are: what kinds of group makeups and individual attributes encourage members to speak their minds, and what determines whether they are heard by the other members? When do group discussions create lasting change, and when do they coerce individuals instead into making decisions that they will regret?

With the rapid advance of Large Language Models, it is now possible to assign detailed traits to AI agents that simulate the properties of diverse individuals, and have them engage in natural language discourse reminiscent of the studies on human collective decision-making \citep{brown2019group}. If perfected, such simulations can provide distinct benefits for behavioral and computational sciences.

Human studies with groups are expensive and difficult to conduct, as they involve coordinating the efforts and schedules of several individuals. Attempting to balance groups with certain identities (e.g., nationality when the decisions are about foreign policy), makes the data gathering even more complicated. If AI agents can adequately mimic the behaviors of individuals with certain attributes who engage in a group discussion, multi-agent systems can serve as a fertile test bed for developing and evaluating hypotheses about functional or dysfunctional collective decisions. Beyond the benefits for the basic sciences, such simulations can be helpful in applied domains where the sensitive nature of the topic prevents large-scale human experiments. One example is modeling the potential outcomes of negotiations to aid a diplomatic team in planning their approach before fateful meetings begin. Another is predicting the course of a proposed bill through a legislature and the kinds of criticisms it would attract from the opposition, so that the most prudent version of the bill is advanced. However, for these scientific and applied benefits of AI discourse simulations to come to fruition, the ability of the agents to accurately and reliably mimic humans in the assigned roles must be confirmed: An agent that is prone to easily changing its stance with or without opposition would not serve as a good model of international negotiators or congresspeople. Even more egregious for such applications would be cases where the very personality that is assigned (nationality in the example above), is abandoned somewhere in the discussion. If such problems are absent, then the dynamics of the discourse must still match those observed in humans for the simulations to hold predictive value. For instance, if human experiments indicate that the pressure to conform to the majority greatly subsides with even a single co-dissident voice \citep{asch1956studies}, such behavior should also emerge within AI simulations.

Studying discourse dynamics in multi-agent systems can also benefit the artificial intelligence community. It is now well-attested that deep-seated sociocultural biases pervade the behaviors of major LLMs \cite{deshpande-etal-2023-toxicity, salewski2024context}, proving difficult to root out even with human-guided debiasing \cite{hemmatian2023muslim}. An example of an undesirable bias is the recorded tendency of the models to present a West-centric view on issues the understanding of which benefits from a global pespective \cite{ varshney2023decolonial, naous2023having}. Because LLMs are often used for diverse tasks without additional fine-tuning, such biases contaminate every downstream output and cause real harms (\citealt{abbasi2019fairness,varshney2019pretrained}). One potential way to mitigate such harms is to incorporate an intermediate diverse group deliberation step before the AI's response is passed on to the human user. If AI agents in a group chat can accurately and reliably represent the identities assigned (e.g., Syrian versus American individuals when discussing the Syrian Civil War), the group's consensus decision is more likely to reflect the nuances of the issue and the opinions of marginalized identities. That is, to the extent that the discourse dynamics also reflect those of humans in more diverse groups ~\citep{sulik2022diversity}\footnote{Note that diversity in group identities and opinions is not always positive. If an extremist person joins a group discussion, we would want the other members' opinions to be highly resistant to their persuasion attempts. However, in many cases, it is possible to exclude such extreme members through screening before the group discussion \citep{brown2019group}. Our mentions of diversity of opinion in the rest of this article in a positive light is conditioned on such a screening process.}.

Although prior work on multi-agent collaboration has demonstrated its benefits in applications such as mathematical reasoning~\citep{du2023improving}, code generation~\citep{hong2024metagpt} and common sense reasoning~\citep{xiong2023examining}, the stability and quality of discourse dynamics remain largely unstudied. It is particularly important to fill this gap in cultural domains, as cultural personas tend to be more complex, less explicit in natural language, and subject to widespread model biases \cite{deshpande-etal-2023-toxicity, salewski2024context}. Our study aims to clarify AI discourse dynamics in one such domain in order to enable the kinds of applications mentioned above. 

We specifically examine the ability of OpenAI's GPT-3.5-Turbo model to simulate inter-cultural collaboration and debate using an experimental framework grounded in large-scale polls about international relations opinions~\citep{durmus2023towards}. Using pre- and post-discussion private responses in conjunction with multi-agent chat transcripts, we test the stability of national personas and their individual opinions as well as the effects of either on group outcomes.\footnote{Code is available at \href{https://github.com/baltaci-r/CulturedAgents}{https://github.com/baltaci-r/CulturedAgents}}

\indent To preview, we find multi-agent discussions to be effective in producing collective decisions that more often reflect diverse perspectives. The benefits, however, are reduced by the AI agents' susceptibility to conformity during discussions, along with their imperfect ability to maintain consistent personas and opinions. These problems persist with instructions that emphasize debating in support of one's opinion rather than collaboration with the other agents. Our results have implications for the use of multi-agent frameworks to reduce cultural bias in LLMs. The mere inclusion of diverse personas may not mitigate biases unless the sources of instability in their contributions, particularly conformity, are addressed. Addressing such issues would also enhance the quality of AI simulations for predicting the outcomes of scientific studies,  negotiations, legislative sessions and wargaming \citep{hua2023war}, contexts that rely heavily on consistent personas for realism. As such, our work motivates further studies on how the constancy of AI personas can be improved. 

\section{Background}

\indent Multi-agent collaboration frameworks draw inspiration from collaborative teamwork observed in human settings. In these frameworks, multiple instances of language models are employed within a cooperative environment to accomplish a complex task \citep{li2024camel,chen2023agentverse}. Collaborative behaviors in humans such as team dynamics and cohesion, leadership, and communication have been thoroughly studied in the human sciences (e.g., \citealt{gupta2022transactive}). In contrast, few studies have examined behaviors in multi-agent language model systems. \citet{li2023theory} observed evidence of emergent collaborative behaviors and high-order Theory of Mind capabilities among LLM-based agents. But \citet{xiong2023examining} highlighted several consistency concerns in multi-agent collaboration, including agents compromising with the opponents and easily changing perspective in a debate, particularly when weaker models interact with superior LLMs. \citet{zhang2023exploring} placed agents in entirely homogeneous groups in terms of thinking patterns and compared the results to settings where one agent exhibits a different thinking approach. They noted the tendency of LLM agents to produce human-like social behaviors in these contexts, such as conformity due to perceived peer pressure. However, the multi-agent societies composed of agents with different traits did not clearly differ in performance. 

Prior research on collaborative behaviors in multi-agent LLM systems has been entirely focused on domains like mathematical reasoning where clear gold standard answers exist, rather than topics like politics where the constancy of personas and viewpoints is more important for faithfully simulating the real world and conflicting views may have complementary value. To address this gap, we study culture-sensitive AI ensembles using the GlobalOpinionQA, a dataset of cross-national surveys gathering diverse opinions on global issues across countries~\citep{durmus2023towards}. 
We assign AI agents with different national personas to groups of five, where they provide initial responses to a question privately before engaging in a peer-moderated discussion about it with the other agents. Once the group discussion is terminated and a collective response is determined, we ask each agent about its opinions on the issue in private once more. 

We focus our analysis on three situations where persona inconstancy is arguably rarely desirable. When agents express an opinion in line with their teammates during conversation that differs from both their pre- and post-discussion response, we are faced with AI behavior that resembles \emph{conformity} due to peer pressure as studied in humans~\citep{asch1956studies,brandstetter2014peer}. A type of inconstancy more closely resembling \emph{confabulation} in clinical conditions arises when the post-discussion opinion does not match either the pre-discussion response or any of the positions proposed during discussion \citep{schacter1995memory}. The third type of inconstancy emerges when an agent instructed to represent a given national identity ''role-plays'' a different persona simply because it was mentioned in discussion, arguably similar to \emph{impersonation} behaviors in individuals with Antisocial Personality Disorder \citep{padhye2012virtual}. 

By systematically manipulating the degree of disagreement within groups (measured using entropy states), we explore whether the frequency of conformity changes as a function of opinion popularity, a key factor in the emergence of similar actions in humans. To test whether encouraging a debate rather than a collaboration environment would induce greater constancy in personas, we look at discourse outcomes across entropy states for both types of interaction.

\section{Experiments}

We use GPT-3.5-turbo with AutoGen, an open-source framework for multi-agent collaboration~\citep{wu2023autogen}. Our experimental setup follows a three-step process (see  Appendix \ref{personna.instability} for a complete example). During the \textit{Onboarding} phase, AI agents are instructed to adopt the national personas present in the dataset for a given question and asked to respond to it in isolation. Agents' responses are compared to the human survey distributions using a cross entropy loss. Agents whose responses do not align with the assigned persona are excluded. 
The diversity of opinions within a group is measured using Shannon entropy, applied to the opinions of agents during Onboarding. This is calculated as $S=-\sum_{o \in \mathcal{B}}{p(o)\log{p(o)}}$, where $p(o)$ represents the relative frequency of the unique opinion $o$ in the set $\mathcal{B}$ of agent responses at Onboarding. Seven entropy classes are obtained for a selection of five agents with the lowest entropy class corresponding to five agents with the same opinion and the highest entropy class with every agent presenting a unique response (see Table~\ref{tab:debate-peer-pressure-and-influence}). To obtain a balanced distribution of different entropy levels across all discussion groups, agent combinations corresponding to the least represented entropy class are chosen for each question as illustrated in Appendix \ref{agent.selection}. Each debate or collaborative discussion is moderated by a chat manager who selects the order of agents for responding to the given question. Discussion is terminated when any agent requests it to be. The chat manager then summarizes the discussion and reports the group's final opinion. The agents then undergo a final \textit{Reflection} step where an assistant agent interviews them to answer the same question one last time independently and privately. 

Based on human research~\citep{asch1956studies,brandstetter2014peer}, we focus our \textit{conformity} analysis on the following entropy levels expected to show peer pressure to different degrees: $4 \oplus 1$ (\textit{lone dissenter}),  $3 \oplus 2$ (\textit{close call}), $3 \oplus 1 \oplus 1$ (\textit{split opposition}). Prior work has shown that even one additional person supporting the less popular view greatly reduces the pressure to conform. As such, we anticipated the rate of conformity to be highest in the \textit{lone dissenter} and \textit{split opposition} entropy classes. In contrast, we examine the rates of \textit{confabulation} by comparing opinions during reflection with Onboarding and intermediate opinions and  \textit{impersonation} using regular expressions across all entropy classes.



\section{Results}

\subsection{General Effects of Diversity}

\begin{figure}[!h]
\includegraphics[width=\columnwidth]{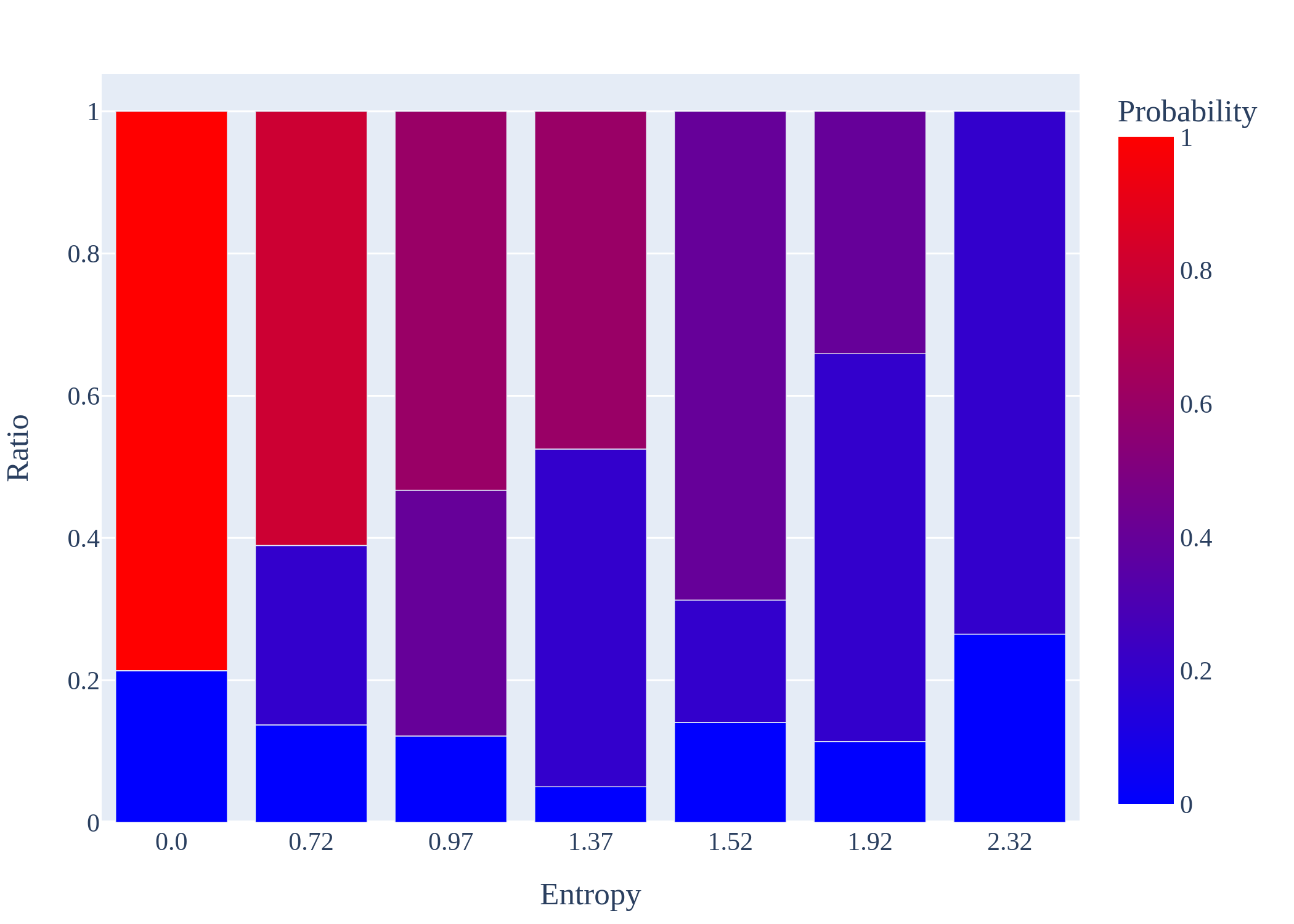}
\caption{Group Prediction follows the distribution of opinions during onboarding  across different onboarding entropy groups for debate while also generating new ideas particularly at the group of highest diversity. Groups are less likely to predict opinions with higher probability for debate  compared to collaboration.}
\label{fig:debate-opinion-entropy_vs_onboarding_probability}
\end{figure}

\indent We first consider the impact of the diversity of agents' opinions during Onboarding on the final group predictions. We measure the ratio of questions in each entropy class with a group response $G$ of relative frequency $p(G)$ as shown for the debate condition in Fig.~\ref{fig:debate-opinion-entropy_vs_onboarding_probability}. We observe that group response largely follows the distribution of opinions during Onboarding across different entropy levels, but it also allows for the generation of new responses regardless of entropy class, particularly for the group with the highest opinion diversity. The same holds for collaboration as displayed in Fig.~\ref{fig:collab-opinion-entropy_vs_onboarding_probability}.

\begin{figure}[h]
\includegraphics[width=\columnwidth]{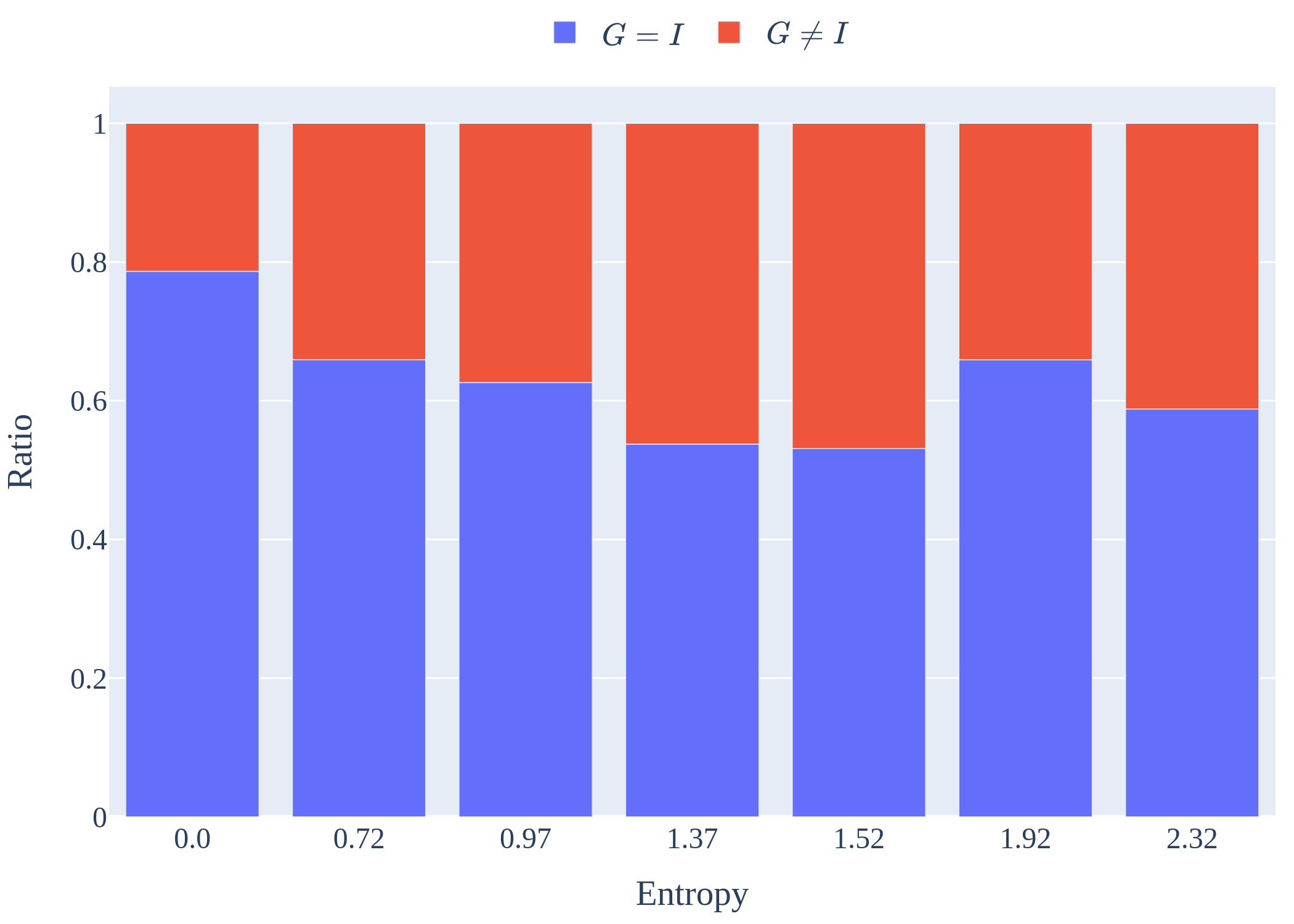}
\caption{Initiators Dominate Group Prediction: agents follow the initiator opinion  of a debate and often converge to the opinion of the initiator $I$. Initiators have less impact on a group's response $G$ during debates compared to collaborations.}
\label{fig:debate-opinion-pred_same_as_initiator}
\end{figure}

However, not all agents have the same degree of influence on group outcomes. The initiator of a discussion has an outsize impact on the group's final decision, regardless of entropy class and even when debate in support of one's position is emphasized for the agents in the instructions (see Fig.~\ref{fig:debate-opinion-pred_same_as_initiator} and Fig.~\ref{collab-opinion-pred_same_as_initiator}). Perhaps unsurprisingly, this influence decreases with increasing diversity of opinions within the group. 

\begin{figure}[h]
\includegraphics[width=\columnwidth]{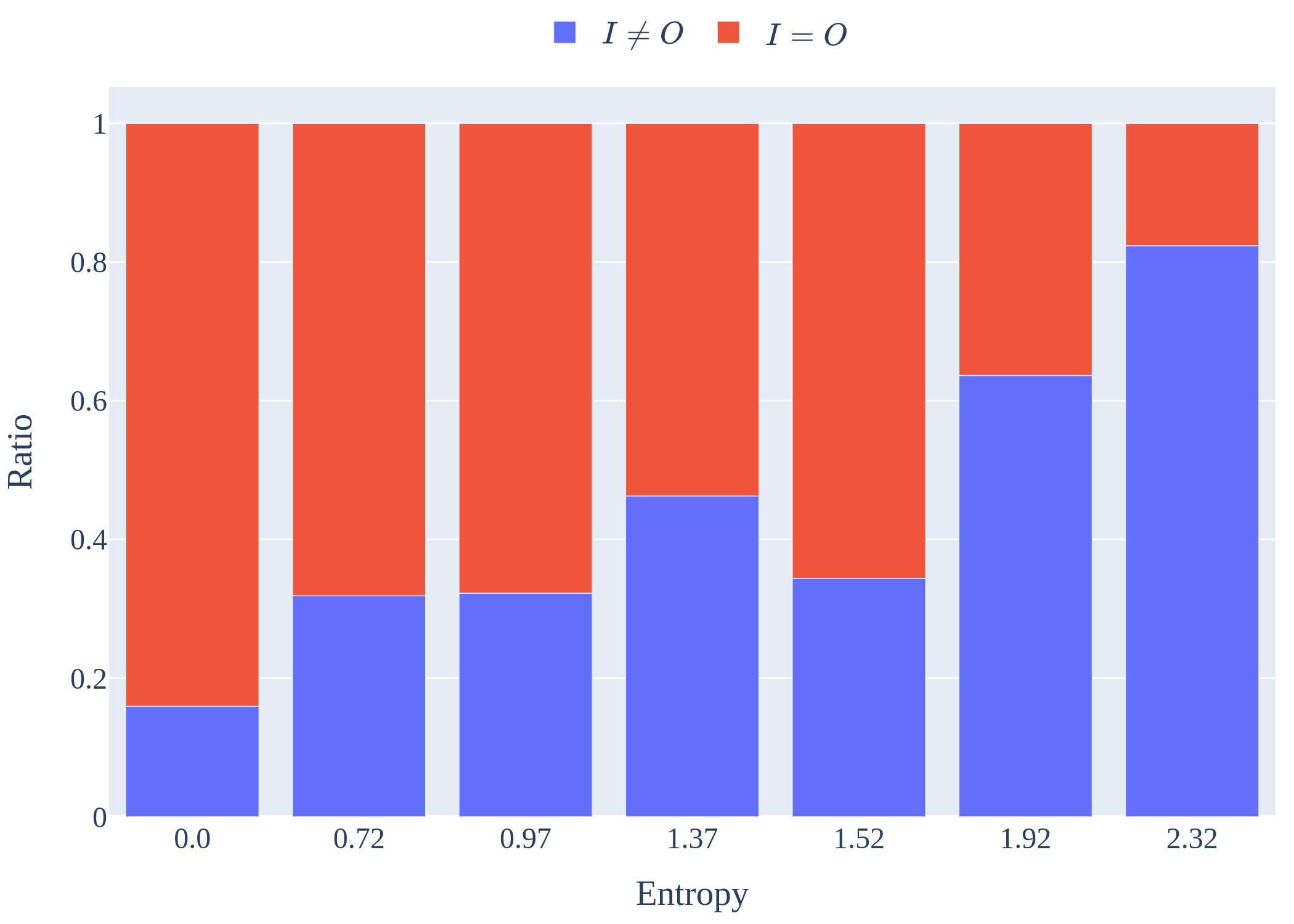}
\caption{The changes in initiator opinion from Onboarding to the onset of a debate can be predicted from the Onboarding entropy of the group opinions. Initiators are more likely to change their opinion as diversity of the group increases despite not having observed the opinions of other agents yet. Initiators of a debate change their opinion during debates less often than in collaborations, highlighting the importance of prompt engineering for inducing persona constancy.}
\label{fig:debate-opinion-initiator_changes_opinion}
\end{figure}

Nonetheless, initiators with minority opinions during Onboarding do not always take advantage of their outsize influence, as they tend to change their expressed views during discussion based on \textit{a priori} perceptions of group opinions. The mere mention of the identities of the debate participants pushes the initiator to change their opinion even before others have spoken (see Fig.~\ref{fig:debate-opinion-initiator_changes_opinion}). As this inconstancy is precipitated by opposing views of interlocutors, it can be characterized as \textit{conformity} due to perceived peer pressure. The dynamics, however, are somewhat different from what is observed in humans, as \textit{any} opinion with a supporter seems to exert an influence regardless of its dominance or lack thereof within the group ~\citep{asch1956studies,brandstetter2014peer}. A similar pattern is observed for collaboration as displayed in Fig.~\ref{fig:collab-opinion-initiator_changes_opinion}.

\indent We further investigate the impact of group diversity on the opinions of individual agents upon reflection. We measure the percentage of agents with opinions of Onboarding probability $p(o)$ that change opinion during the Reflection phase compared to agents that keep their opinion. We also measure the average ratio of showing a different intermediate response compared to the Reflection opinions for individual agents. We further compare the percentage of agents with an opinion corresponding to the group response compared to agents with a different reflection opinion. We are particularly interested in dominated agents as shown in bold under the Group column in Tab.~\ref{tab:debate-peer-pressure-and-influence}, as they are most important for diverse outcomes in real life settings. 

\textit{Lone dissenters} ($S = 0.72$) are the most likely to change their opinions upon reflection to align with the group response. When they retain their onboarding positions, they present a different viewpoint during the discussion about half of the time. Both patterns align well with the findings from classic research on conformity due to peer pressure ~\citep{asch1956studies,brandstetter2014peer}. As expected, dominated agents are comparatively more likely to hold onto their opinions in the \textit{close call} configuration ($S = 0.97$). However, unlike human studies where the presence of a co-dissenter largely eliminates conformity, the minority agents still convert to the majority viewpoint for Reflection most of the time, demonstrating peer influence. Also unlike human studies is the fact that the agents are most receptive to altering their opinions at states of higher entropy ($S=2.32$). Any opinion voiced during Discussion appears to exert an influence on the Reflection stage change of mind, regardless of dominance relationships between the viewpoints. This is in contrast with human studies where dominance relationships between the various opinion strongly influence behavior~\citep{asch1956studies,brandstetter2014peer}.

To summarize, while the peer pressure and influence phenomena observed in our simulations are human-like, their dynamics based on group composition differ significantly from human studies, where \textit{lone dissenter} and \textit{split opposition} dominated agents are most likely to show both peer influence and peer pressure in their decision-making.

\subsection{Inconstant Personas}

\indent In addition to studying the dynamics of group interactions, we identify two rare but highly disruptive forms of persona inconstancy that can negatively impact the quality of complex reasoning in cultural multi-agent systems. One is the agents' tendency to adopt a different persona motivated by prior Discussion context, particularly in the case of debates. Using a simple heuristic to find instances when an agent says ''As an X agent'' where X is incompatible with their assigned national identity, we find that agents adopt a different persona on average once in every 200 messages during a collaborative Discussion session. Debate instructions that explicitly encouraged the agents to stand firm in their beliefs and maintain their personas made such \textit{impersonation} behaviors much rarer ($0.018\%$). 

Another form of inconstancy is an agent's tendency to report an opinion not seen during either Onboarding or Discussion, mimicking the \textit{confabulation} of novel content observed in certain clinical conditions. We find that $1.1\%$ of the opinions at reflection come neither from onboarding nor from the debate statements of any agent. Collaboration conditions showed higher rates of confabulation ($1.64\%$).

%

\begin{table*}[!h]
\centering
\caption{Peer Pressure and Peer Influence in Debate: {Lone dissenters} ($S = 0.72$) are the most likely to change their opinions upon Reflection to align with the group response. When they retain their Onboarding positions, they present a different viewpoint during the Discussion about half of the time. Both patterns are indicative of peer pressure. Dominated agents are comparatively more likely to hold onto their opinions in the $S = 0.97$ entropy class, but still convert to the majority viewpoint for Reflection about half of the time, demonstrating peer influence. The agents are most receptive to altering their opinions at states of higher entropy (e.g., $S=2.32$). Any opinion voiced during Discussion appears to exert an influence on the Reflection stage change of mind, regardless of dominance relationships between the viewpoints. }
\label{tab:debate-peer-pressure-and-influence}
\resizebox{\textwidth}{!}{\begin{tabular}{ccc|cccc|cccc|c}
                     &                                                           & \multicolumn{1}{c}{}       & \multicolumn{4}{c}{$R=O$}                                                      & \multicolumn{4}{c}{$R \neq O$}                                                 &        \\
$S$                  & Group                                                     & \multicolumn{1}{l}{$p(o)$} & $\%$              & $D \neq R$    & $R \neq G$     & \multicolumn{1}{l}{$R=G$} & $\%$              & $D \neq R$    & $R \neq G$     & \multicolumn{1}{l}{$R=G$} & N      \\ 
\hline\hline
$0.0$                & $5$                                                       & $1.0$                      & $\textbf{85.47}^*$ & $0.84$          & $13.12$          & $86.88$                     & $14.53$             & $0.68$          & $40.22$          & $59.78 $                    & $389$  \\ 
\hline
$0.72$               & $4 \oplus \textbf{1}$                                     & $0.2$                      & $54.05 $            & $\textbf{0.73}$ & $53.33$          & $46.67$                     & $45.95$             & $\textbf{0.81}$ & $11.76$          & $\textbf{88.24}$            & $226$  \\
\multicolumn{1}{l}{} & \multicolumn{1}{l}{}                                      & $0.8$                      & $68.74 $            & $0.77$          & $22.0$           & $78.0$                      & $31.26$             & $0.69$          & $32.13$          & $67.87$                     &        \\ 
\hline
$0.97$               & $3 \oplus \textbf{2}$                                     & \textbf{$0.4$}             & $\textbf{56.46}$    & $0.72$          & $46.61$          & $\textbf{53.39}$            & $43.54$             & $0.76$          & $20.88$          & $79.12$                     & $214$  \\
\multicolumn{1}{l}{} & \multicolumn{1}{l}{}                                      & $0.6$                      & $68.7$              & $0.79$          & $31.78$          & $68.22$                     & $31.3$              & $0.76$          & $30.77$          & $69.23$                     &        \\ 
\hline
$1.37$               & $3 \oplus \textbf{1} \oplus \textbf{1}$                   & $0.2$                      & $45.45$             & $0.61$          & $55.71$          & $44.29$                     & $54.55$             & $0.73$          & $23.81$          & $76.19$                     & $80$   \\
\multicolumn{1}{l}{} & \multicolumn{1}{l}{}                                      & $0.6$                      & $57.63$             & $0.69$          & $28.68$          & $71.32$                     & $42.37$             & $0.60$           & $38.0$           & $62.0$                      &        \\ 
\hline
$1.52$               & $2 \oplus 2 \oplus \textbf{1}$                            & $0.2$                      & $36.67$             & $0.72$          & $\textbf{59.09}$ & $40.91$                     & $63.33$             & $0.63$          & $26.32$          & $73.68$                     & $64$   \\
\multicolumn{1}{l}{} & \multicolumn{1}{l}{}                                      & $0.4$                      & $49.79$             & $0.64$          & $45.45$          & $54.55$                     & $50.21$             & $0.66$          & $33.61$          & $66.39$                     &        \\ 
\hline
$1.92$               & $2 \oplus \textbf{1} \oplus \textbf{1} \oplus \textbf{1}$ & $0.2$                      & $35.38$             & $0.55$          & $50.0$           & $50.0$                      & $\textbf{64.62}$    & $0.68$          & $\textbf{29.76}$ & $70.24$                     & $44$   \\
\multicolumn{1}{l}{} & \multicolumn{1}{l}{}                                      & $0.4$                      & $40.91$             & $0.7$           & $38.89$          & $61.11$                     & $59.09$             & $0.64$          & $30.77$          & $69.23$                     &        \\ 
\hline
$2.32$               & $1 \oplus 1 \oplus 1 \oplus 1 \oplus 1$                   & $0.2$                      & $25.61$             & $0.45$          & $59.52$          & $40.48$                     & $\textbf{74.39}^*$ & $0.68$          & $30.33$          & $69.67$                     & $34$  
\end{tabular}}
\end{table*}

\section{Discussion}

We found evidence of sophisticated interaction dynamics in a multi-agent framework for GPT-3.5-Turbo personas with different nationalities that discussed contentious international relations topics. Novel responses emerged from discussions even among entirely homogeneous groups, highlighting the generative nature of multi-agent LLM frameworks. However, a group's initial opinion diversity, the entropy $S$ of private responses during the \textit{Onboarding} stage before inter-agent discussion, emerged as a stronger determinant of conversation contents and collective decisions. This happened regardless of whether the agents were instructed to debate in support of their beliefs or asked to collaborate in the service of collective decision-making.

Opinion diversity seems to exert its effect partly by reducing the outsize influence of chat initiators on collective decisions, but also by causing them to change the views they voice to conform to other agents. Similar to human studies, many agents reverted back to their original opinions when asked about the topic in private after the discussion, identifying their in-chat proclamations as the results of conformity rather than genuine opinion adjustment. That only mentioning the identities of co-interlocutors is sufficient to change the initiator's proposed stance during Discussion speaks to the profound susceptibility of LLMs to peer pressure. 

Some of the dynamics in LLM agents resembled those observed in human studies on peer pressure and influence ~\citep{asch1956studies,brandstetter2014peer}. Being a lone dissenter was the strongest predictor of changing one's opinion during Reflection to align with the majority stance. However, the AI agents' submission to peer pressure or influence was less affected by the dominance relationships between various viewpoints. For instance, a single co-dissenter greatly reduces the rates of a human agent with a minority opinion succumbing to peer pressure ~\citep{asch1956studies}. However, no such mitigation was observed in the \textit{close call} condition where two agents disagreed with the three-person majority view. In fact, as demonstrated in Tables 1 and 2, it seems that any opinion voiced during discussion exerts influence on the Reflection responses in close correspondence to its frequency. One explanation for the difference is a lack of a clear separation between role identities and the linguistic context of the chat for AI agents, unlike human conversations. The co-interlocutors are simply parts of the prompt context for the AI model and their expressed views may therefore activate associated portions of the models' trained weights in close approximation of their frequency.

\indent Unlike \textit{conformity}, which is a normal response to group interactions in humans, other sources of inconstancy more closely resembled abnormal behaviors such as \textit{impersonation} in Antisocial Personality Disorder \citep{padhye2012virtual} and \textit{confabulation} in memory disorders \citep{schacter1995memory}. Our simple heuristic showed that in one out of every 2000 interactions during debate, the agents presented themselves as belonging to a different nationality than the one assigned to them. This was most often a direct reaction to a nationality being mentioned in the last response that was not represented in the group, highlighting the prominence of chat context over role prompting in determining model generations. It is comparatively more difficult to identify the source of \textit{confabulations}, where the models presented opinions during reflection that were neither represented in the chat nor indicated as their pre-discussion response, therefore being completely absent from the linguistic context. These behaviors may reflect the difficulty of maintaining role prompt personas in the face of lengthy chat contexts, or simply the stochastic nature of the LLM responses. Regardless of their source, such unpredictable responses undermine the usability of multi-agent discussions for simulation and debiasing purposes, marking them as important targets for future studies. 

\section*{Limitations}
One limitation of this work is the uneven distribution of examples across entropy classes. This was driven by the unequal representation of global perspectives in the GlobalOpinionsQA dataset \citep{durmus2023towards}, which results in fewer examples for higher entropy classes. We addressed this imbalance by selecting the least represented entropy configuration for each question. Future research should confirm the findings in more balanced datasets. Another limitation arises from the occasional errors of assistant agents in summarizing intermediate replies and generating the collective responses. To enhance the quality of the summarization, we included the options for each question in the associated prompt. But human aggregation of opinions in future research would be helpful to confirm the results. Finally, there were far more patterns in the behaviors of the agents than the handful of phenomena we have highlighted herein. Future work should further explore all the complex and sometimes nonsensical ways in which the AI personas interact. 

\section{Conclusion}

Two major use cases can be imagined for multi-agent AI systems with diverse sociocultural identities. One is to simulate the possible outcomes of group interactions where data from humans is difficult to come by (e.g., behavioral experiments) or is precluded given the sensitive nature of the topic (e.g., for negotiation or legislation modeling). The other is to implement an intermediate diverse group discussion before an LLM's sociocultural responses are provided to a human user, with the hope that it can enhance the nuance and cultural perceptiveness of the final response. However, both of these use cases crucially depend on the AI agents maintaining their assigned identities and persevering in their beliefs in the absence of strong arguments to the contrary. We found evidence of current multi-agent systems occasionally struggling with this constancy. 

Culture-sensitive AI agents are susceptible to peer influence and pressure even as chat initiators. This highlights the importance of studying conversational dynamics in multi-agent systems, rather than taking the ''collective decision'' outcomes of group discussions at face value. The examination of such dynamics is particularly important for cultural issues: The mere inclusion of a minoritized identity in group chats does not necessarily translate into less biased discussion outcomes if the minoritized agent does not voice its opinion freely or reliably. Fortunately, our results suggest private post-discussion interrogations of models can counteract some of the pressure produced by the majority opinion, similar to what has been found in human conformity experiments \citep{asch1956studies}. This provides a way to make outputs drawn from multi-agent frameworks more representative of diverse perspectives. 

Work on understanding multi-agent dynamics will also need to incorporate measures of persona and response constancy. Agents come up with post-discussion responses that do not arise naturally from either the assigned personas or the discussion content. In some cases they even drop the personas altogether to impersonate a completely different, absent national identity. Such sources of irrational responding would cast serious doubt on the results of multi-agent systems' reasoning if not properly measured and addressed. 

This encourages further investigations that identify the sources of inconstancy and effective ways to reduce them, so that multi-agent discourse can fulfill its potential in basic and applied scientific disciplines. Accordingly, we are currently exploring prompting and agent-based modeling strategies to reduce these sources of unreliability. We hope this work will encourage further research within the AI community on inter-agent dynamics, particularly for simulations on cultural issues where the debiasing influence of diverse views is needed the most.

\section*{Ethics Statement}

This study explores interactions among simulated national personas in debate and collaboration scenarios. Research indicates that LLMs can generate harmful viewpoints or toxic content during these interactions~\citep{liu2023jailbreaking}. The authors explicitly disapprove of any offensive conduct by the simulated agents. The group discussions presented here are solely for research purposes, aimed at enhancing comprehension of cultured multi-agent systems dynamics. 

\bibliography{custom}
\bibliographystyle{acl_natbib}

\appendix
\clearpage
\section{Dynamics of Collaboration}
\label{sec:appendix}

\begin{figure}[h]
\includegraphics[width=0.9\columnwidth]{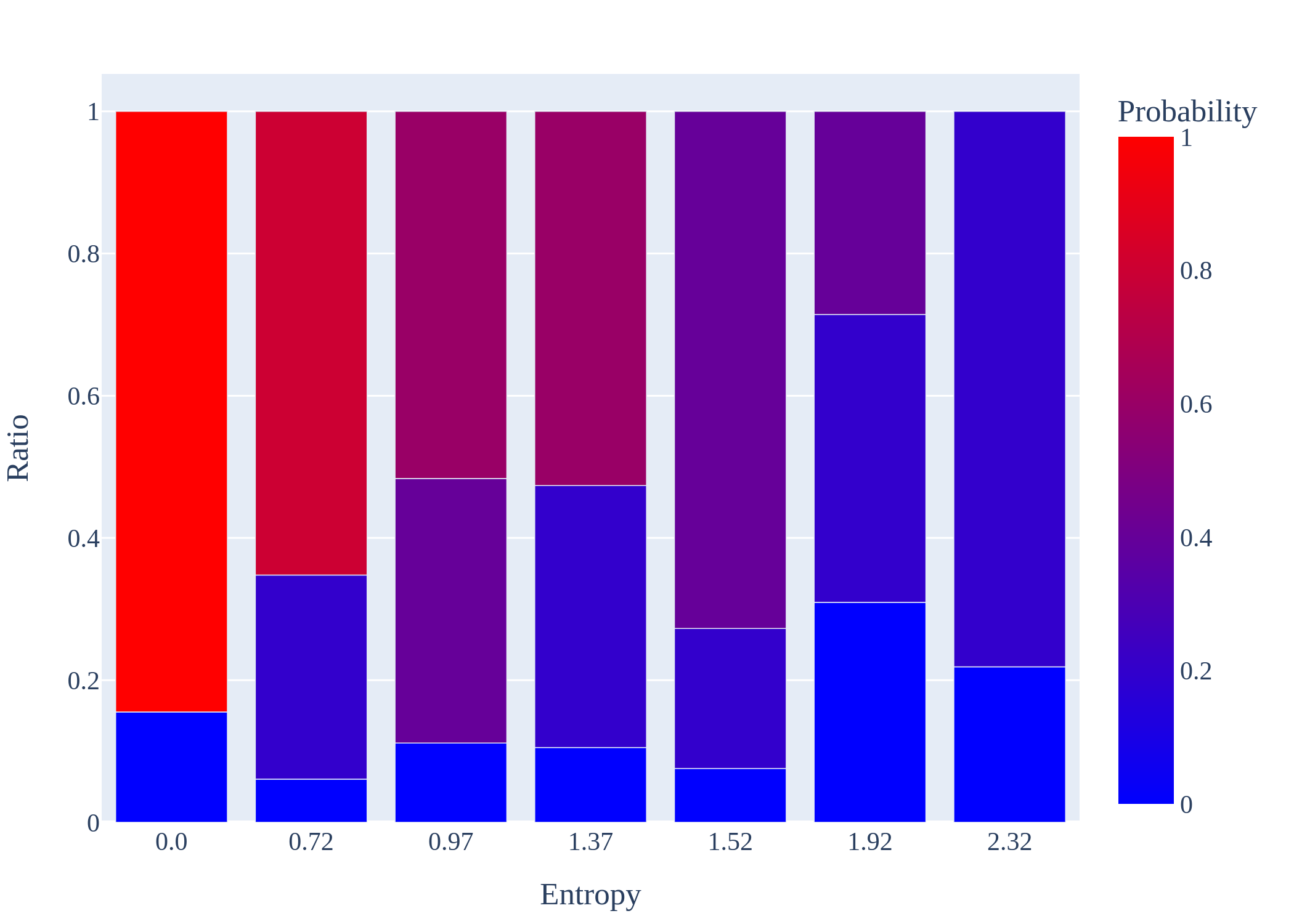}
\caption{Group Prediction in a collaboration follows opinions with higher probabilities across different onboarding entropy groups. Groups are more likely to predict opinions with higher probability for collaboration compared to debate. Generation of new ideas occurs at different entropies particularly at the group of highest diversity.}
\label{fig:collab-opinion-entropy_vs_onboarding_probability}
\end{figure}

\begin{figure}[!h]
\includegraphics[width=0.9\columnwidth]{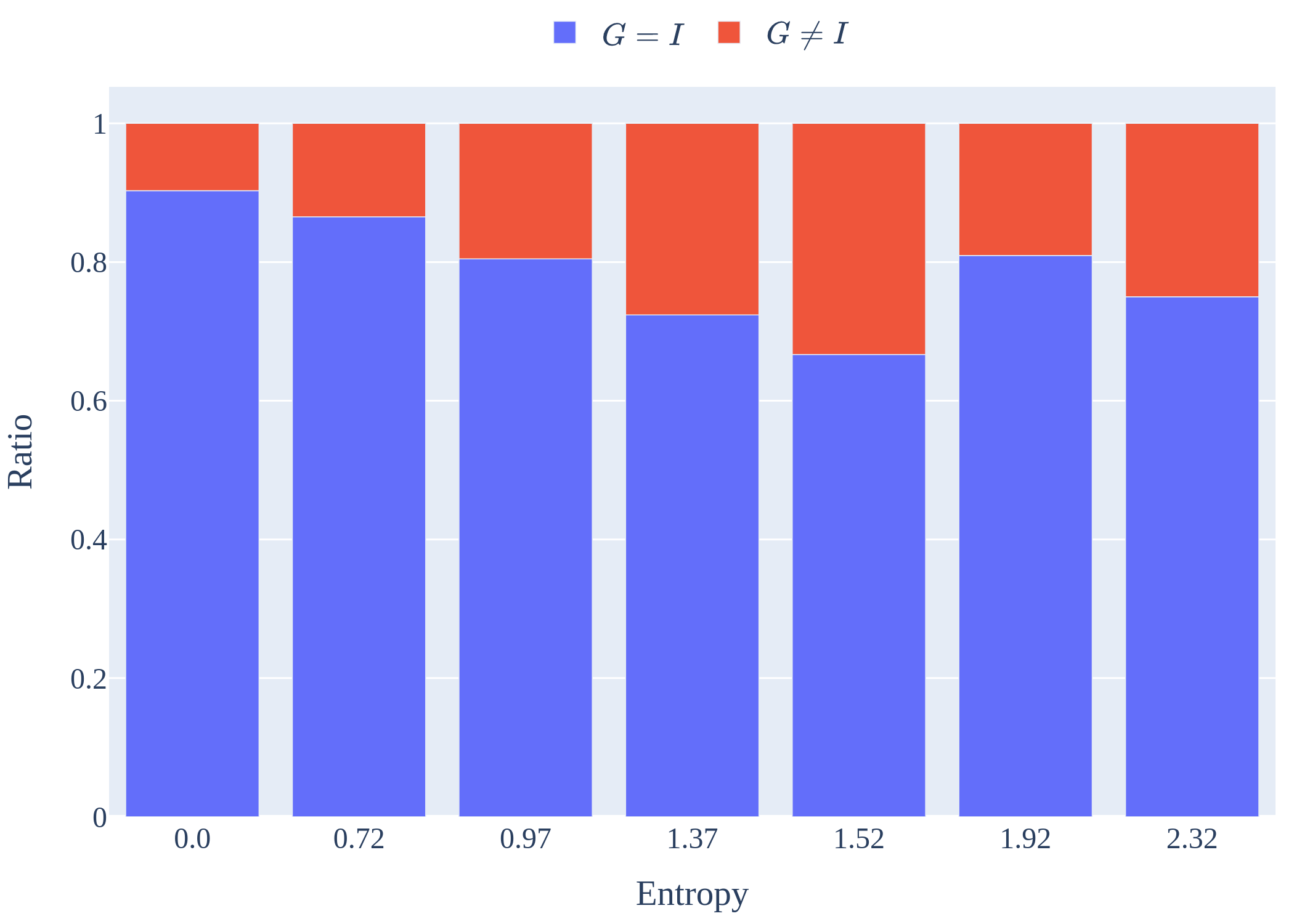}
\caption{Initiators Dominate Group Prediction: agents follow the initiator of a collaboration and often converge to the opinion of the initiator  $I$ as the group prediction $G$.}
\label{collab-opinion-pred_same_as_initiator}
\end{figure}

\begin{figure}[!h]
\includegraphics[width=0.9\columnwidth]{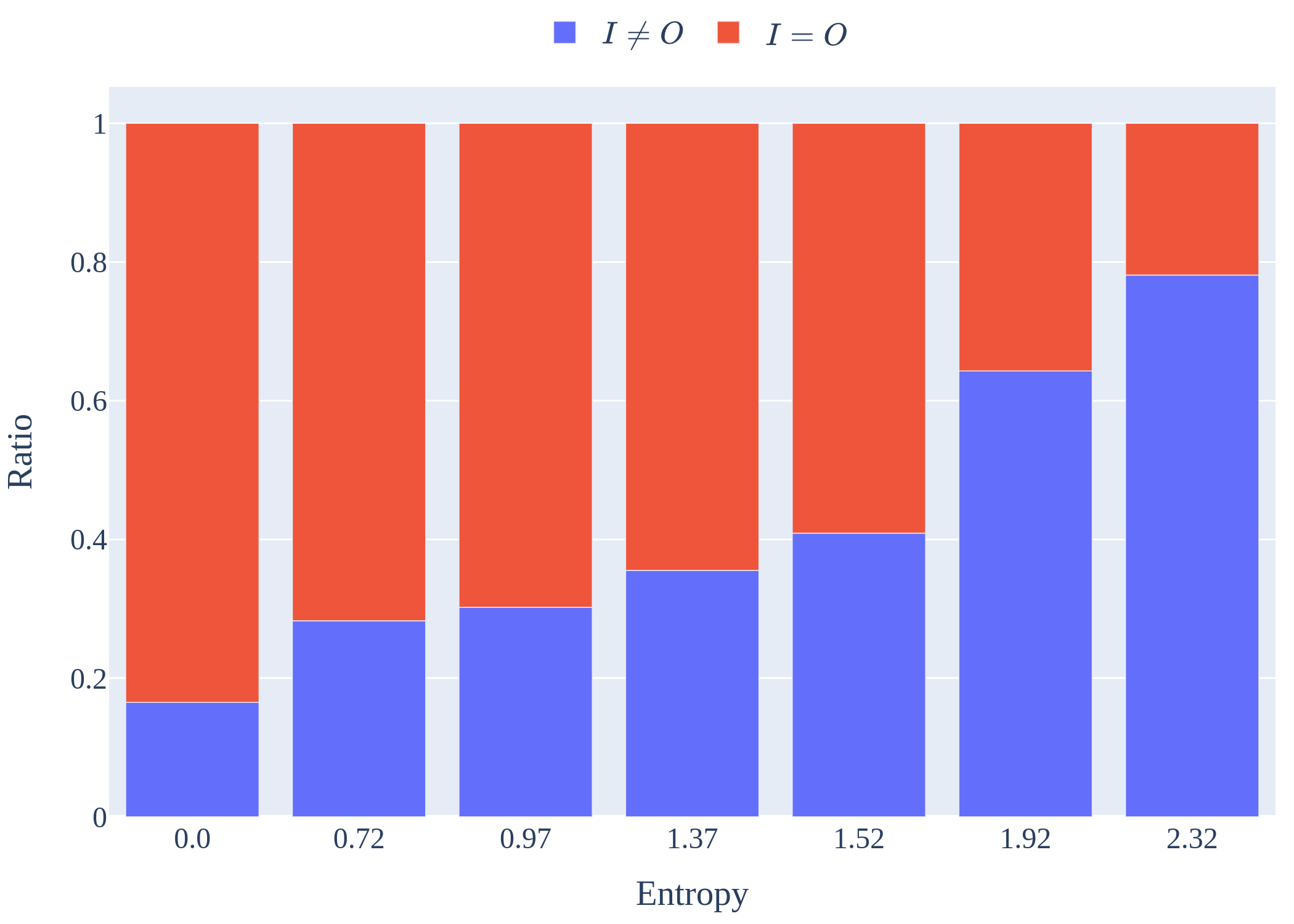}
\caption{Initiator changes its opinion $O$ during onboarding to $I$ at the onset of a collaboration depending on onboarding entropy.} 
\label{fig:collab-opinion-initiator_changes_opinion}
\end{figure}


\begin{figure*}[!ht]
    \centering
    \begin{subfigure}{0.45\textwidth}
        \centering
        \includegraphics[width=\linewidth]{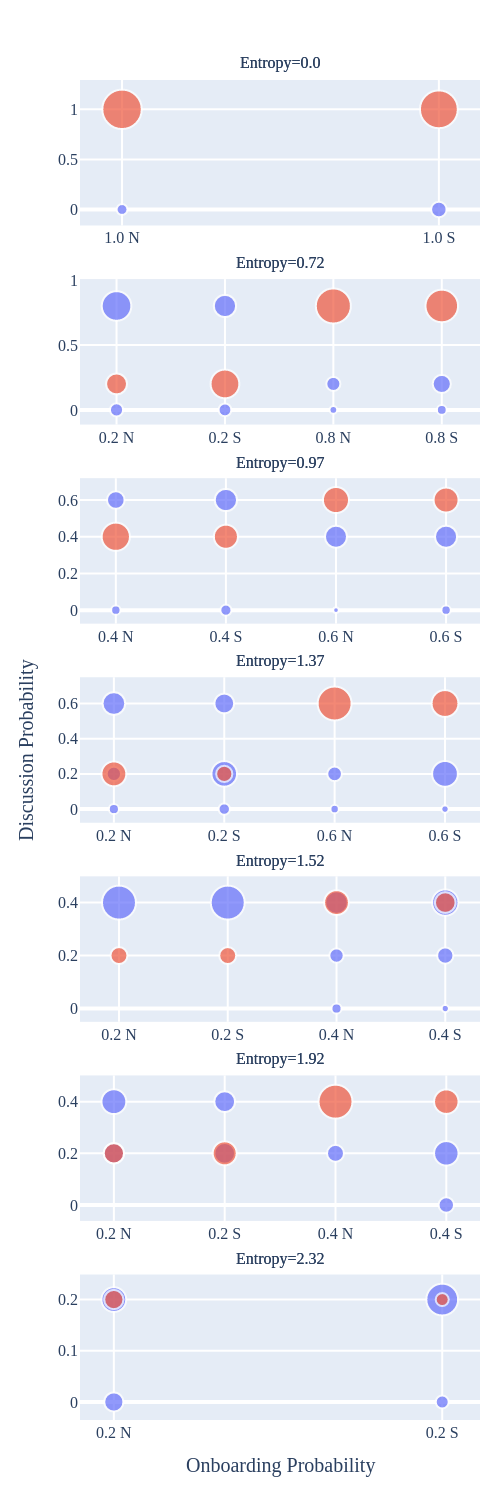}
        \caption{Debate}
        \label{fig:debate-opinion-initiator_changes_opinion_by_region}
    \end{subfigure}
    \hfill
    \begin{subfigure}{0.45\textwidth}
        \centering
        \includegraphics[width=\linewidth]{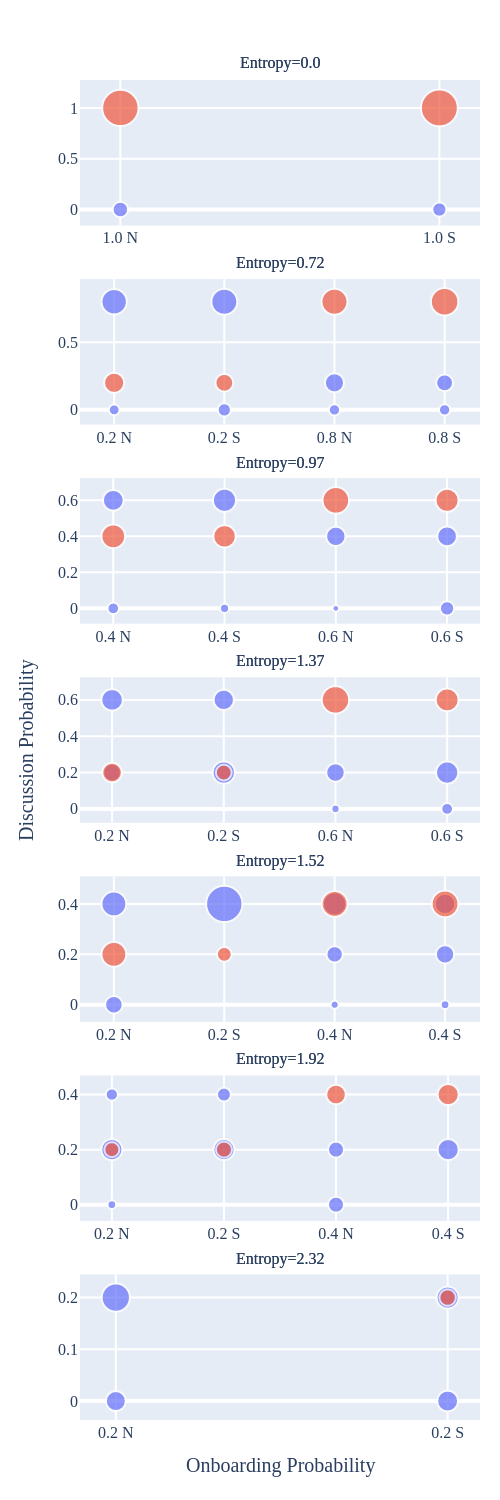}
        \caption{Collaboration}
        \label{fig:collab-opinion-initiator_changes_opinion_by_region}
    \end{subfigure}
    
    \caption{Initiator changes its opinion depending on onboarding entropy and opinion probability provided for Global South $S$ and Global North $N$ nationalities. Initiators change their opinion at the onset of a discussion to match the most probable opinion during onboarding despite not observing agents opinions. Initiators are more likely to change opinion in collaboration compared to debate. }
    \label{opinion-initiator_changes_opinion_by_region}
\end{figure*}

\begin{table*}[!h]
\centering
\caption{Peer Pressure and Peer Influence in Collaboration: Agents maintain their opinions $O$ most strongly in the lowest entropy states during reflection $R$ after debate, while being most open to changing their opinions in the highest entropy state. When dominated in discussions, agents are most resistant to opinion change during reflection in \textit{close call} configuration ($S = 0.97$) and most susceptible to change in high entropy states ($S = 1.92$). During discussions, agents express intermediate discussion opinions $D$ most contrary to their reflection and onboarding opinions in the relatively high entropy configuration ($S = 1.52$), indicating high peer pressure. Dominated agents exhibit the highest peer influence by following group predictions during opinion changes in higher entropy states ($S = 1.92$).}
\resizebox{\textwidth}{!}{\begin{tabular}{ccc|cccc|cccc|c}
                     &                                                           & \multicolumn{1}{l}{}       & \multicolumn{4}{c}{$R=O$}                                                           & \multicolumn{4}{c}{$R \neq O$}                                                      & \multicolumn{1}{l}{}  \\
$S$                  & Group                                                     & \multicolumn{1}{l}{$p(o)$} & $\%$               & $D \neq R$      & $R \neq G$       & \multicolumn{1}{l}{$R=G$} & $\%$               & $D \neq R$      & $R \neq G$       & \multicolumn{1}{l}{$R=G$} & N                     \\ 
\hline\hline
$0.0$                & $5$                                                       & $1.0$                      & $\textbf{86.97}^*$ & $0.88$          & $6.58$           & $\textbf{93.42}^*$        & $13.03$            & $0.7$           & $30.15$          & $69.85$                   & $412$                 \\ 
\hline
$0.72$               & $4 \oplus \textbf{1}$                                     & $\textbf{0.2}$             & $48.66$            & $0.62$          & $48.62$          & $51.38$                   & $51.34$            & $\textbf{0.82}$ & $11.3$           & $\textbf{88.7}$           & $230$                 \\
\multicolumn{1}{l}{} & \multicolumn{1}{l}{}                                      & $0.8$                      & $72.09$            & $0.81$          & $15.4$           & $84.6$                    & $27.91$            & $0.75$          & $24.1$           & $75.9$                    &                       \\ 
\hline
$0.97$               & $3 \oplus \textbf{2}$                                     & $\textbf{0.4}$             & $\textbf{57.55}$   & $0.65$          & $39.58$          & $\textbf{60.42}$          & $42.45$            & $0.8$           & $15.25$          & $84.75$                   & $215$                 \\
\multicolumn{1}{l}{} & \multicolumn{1}{l}{}                                      & $0.6$                      & $62.05$            & $0.76$          & $24.37$          & $75.63$                   & $37.95$            & $0.78$          & $17.84$          & $82.16$                   &                       \\ 
\hline
$1.37$               & $3 \oplus \textbf{1} \oplus \textbf{1}$                   & $\textbf{0.2}$             & $43.84$            & $0.49$          & $57.81$          & $42.19$                   & $56.16$            & $0.78$          & $17.07$          & $82.93$                   & $76$                  \\
\multicolumn{1}{l}{} & \multicolumn{1}{l}{}                                      & $0.6$                      & $60.45$            & $0.79$          & $21.8$           & $78.2$                    & $39.55$            & $0.6$           & $37.93$          & $62.07$                   &                       \\ 
\hline
$1.52$               & $2 \oplus 2 \oplus \textbf{1}$                            & $\textbf{0.2}$             & $32.31$            & $\textbf{0.69}$ & $47.62$          & $52.38$                   & $67.69$            & $0.7$           & $20.45$          & $79.55$                   & $66$                  \\
\multicolumn{1}{l}{} & \multicolumn{1}{l}{}                                      & $0.4$                      & $54.72$            & $0.62$          & $43.17$          & $56.83$                   & $45.28$            & $0.71$          & $23.48$          & $76.52$                   &                       \\ 
\hline
$1.92$               & $2 \oplus \textbf{1} \oplus \textbf{1} \oplus \textbf{1}$ & $\textbf{0.2}$             & $32.23$            & $0.32$          & $\textbf{66.67}$ & $33.33$                   & $\textbf{67.77}$   & $0.71$          & $\textbf{23.17}$ & $76.83$                   & $42$                  \\
\multicolumn{1}{l}{} & \multicolumn{1}{l}{}                                      & $0.4$                      & $45.57$            & $0.56$          & $47.22$          & $52.78$                   & $54.43$            & $0.59$          & $34.88$          & $65.12$                   &                       \\ 
\hline
$2.32$               & $1 \oplus 1 \oplus 1 \oplus 1 \oplus 1$                   & $0.2$                      & $28.48$            & $0.34$          & $62.79$          & $37.21$                   & $\textbf{71.52}^*$ & $0.68$          & $29.63$          & $70.37$                   & $32$                 
\end{tabular}}
\end{table*}

\clearpage
\section{Persona Inconstancy}
\label{personna.instability}
\subsection{Onboarding:} 
\begin{tcolorbox}[title=\emojirobot \texttt{Assistant Agent}]
 This is an onboarding stage. At this stage, responses remain anonymous and are not shared with other participants. Answer the following multiple-choice question independently: \\
Now I'm going to read you a list of countries.  For each one, please tell me whether you would favor or oppose admitting it into the European Union in the next ten years...Ukraine
\\A. Favor
\\B. Oppose
\end{tcolorbox}

\begin{tcolorbox}[title=\emojiuk \texttt{British Agent} ]
A. Favor \emojicheckmark
\end{tcolorbox}

\begin{tcolorbox}[title=\emojibulgaria \texttt{Bulgarian Agent} ]
A. Favor  \emojicheckmark
\end{tcolorbox}

\begin{tcolorbox}[title=\emojiczechia \texttt{Czech Agent} ]
A. Favor  \emojicheckmark
\end{tcolorbox}

\begin{tcolorbox}[title=\emojifrance
\texttt{French Agent} ]
B. Oppose  \emojicrossmark
\end{tcolorbox}

\begin{tcolorbox}[title=\emojigermany
\texttt{German Agent} ]
A. Favor \emojicrossmark
\end{tcolorbox}

\begin{tcolorbox}[title=\emojihungary
\texttt{Hungarian Agent} ]
B. Oppose \emojicrossmark
\end{tcolorbox}

\begin{tcolorbox}[title=\emojiitaly \texttt{Italian Agent}]
B. Oppose \emojicheckmark 
\end{tcolorbox}

\begin{tcolorbox}[title=\emojilithuania
\texttt{Lithuanian Agent} ]
A. Favor \emojicheckmark
\end{tcolorbox}

\begin{tcolorbox}[title=\emojipoland \texttt{Polish Agent} ] 
A. Favor \emojicheckmark
\end{tcolorbox}

\begin{tcolorbox}[title=\emojislovakia \texttt{Slovak Agent}]
A. Favor \emojicheckmark
\end{tcolorbox}

\begin{tcolorbox}[title=\emojispain \texttt{Spanish Agent}]
A. Favor \emojicheckmark
\end{tcolorbox}

\subsection{Agent Selection:}
\label{agent.selection}
We generate all combinations of agents with representative opinions (\emojicheckmark) compared to ground truth labels as displayed in Tab. \ref{tab:combs}. We select the entropy value that corresponds to the least represented class to maintain a balanced dataset across entropy configurations. For this example, the least represented class compared to other entropy classes with a larger number of examples is  $S=0.72$. We randomly select one combination of \emojibulgaria \emojiitaly \emojipoland \emojislovakia \emojispain. 

\begin{table}[h]
    \centering
    \arrayrulecolor{black}
    \begin{tabular}{|l|c|} 
    \hline
    Combinations                                                                                                                                                                                                                                           & $S$                       \\ 
    \hline
    \emojiitaly \emojilithuania \emojipoland \emojislovakia \emojispain                                                            & \multirow{15}{*}{$0.72$}  \\
    \emojibulgaria \emojiitaly \emojilithuania \emojipoland \emojislovakia                                                         &                             \\
    \emojibulgaria \emojiitaly \emojilithuania \emojipoland \emojispain                                                            &                             \\
    \emojibulgaria \emojiitaly \emojilithuania \emojislovakia \emojispain                                                          &                             \\ 
    \arrayrulecolor{red}\cline{1-1}
    \multicolumn{1}{!{\color{red}\vrule}l!{\color{red}\vrule}}{\emojibulgaria \emojiitaly \emojipoland \emojislovakia \emojispain \emojigamedie} &                             \\ 
    \cline{1-1}
    \emojiczechia \emojiitaly \emojilithuania \emojipoland \emojislovakia                                                          &                             \\
    \emojiczechia \emojiitaly \emojilithuania \emojipoland \emojispain                                                             &                             \\
    \emojiczechia \emojiitaly \emojilithuania \emojislovakia \emojispain                                                           &                             \\
    \emojiczechia \emojiitaly \emojipoland \emojislovakia \emojispain                                                              &                             \\
    \emojibulgaria \emojiczechia \emojiitaly \emojilithuania \emojipoland                                                          &                             \\
    \emojibulgaria \emojiczechia \emojiitaly \emojilithuania \emojislovakia                                                        &                             \\
    \emojibulgaria \emojiczechia \emojiitaly \emojilithuania \emojispain                                                           &                             \\
    \emojibulgaria \emojiczechia \emojiitaly \emojipoland \emojislovakia                                                           &                             \\
    \emojibulgaria \emojiczechia \emojiitaly \emojipoland \emojispain                                                              &                             \\
    \emojibulgaria \emojiczechia \emojiitaly \emojislovakia \emojispain                                                            &                             \\ 
    \arrayrulecolor{black}\hline
    \emojibulgaria \emojiczechia \emojipoland \emojislovakia \emojispain                                                           & \multirow{6}{*}{$0$}      \\
    \emojibulgaria \emojiczechia \emojilithuania \emojislovakia \emojispain                                                        &                             \\
    \emojibulgaria \emojiczechia \emojilithuania \emojipoland \emojispain                                                          &                             \\
    \emojibulgaria \emojiczechia \emojilithuania \emojipoland \emojislovakia                                                       &                             \\
    \emojibulgaria \emojilithuania \emojipoland \emojislovakia \emojispain                                                         &                             \\
    \emojiczechia \emojilithuania \emojipoland \emojislovakia \emojispain                                                          &                             \\
    \hline
    \end{tabular}
    \caption{Combinations of agents with corresponding entropy values. The entropy is calculated as follows, for example, $S($\emojibulgaria \emojiitaly \emojipoland \emojislovakia \emojispain$)=-{p(A)\log{p(A)}}-{p(B)\log{p(B)}}=-{4/5\log{(4/5)}}-{1/5\log{(1/5)}}=0.72$}
    \label{tab:combs}
    \end{table}

\begin{tcolorbox}[title=\emojibulgaria 
\texttt{Bulgarian Agent} ]
A. Favor
\end{tcolorbox}

\begin{tcolorbox}[title=\emojiitaly \texttt{Italian Agent}]
A. Oppose
\end{tcolorbox}

\begin{tcolorbox}[title=\emojislovakia \texttt{Slovak Agent}]
A. Favor
\end{tcolorbox}

\begin{tcolorbox}[title=\emojipoland \texttt{Polish Agent} ]
A. Favor
\end{tcolorbox}

\begin{tcolorbox}[title=\emojispain
\texttt{Spanish Agent} ]
A. Favor
\end{tcolorbox}

\subsection{Debate:}
\begin{tcolorbox}[title=\emojirobot \texttt{Chat Manager}] 
You are participating in a debate with fellow agents. You are tasked with defending your perspective. Take a stance and present compelling arguments to support your position. Engage in respectful discourse with other agents, counter their arguments, and provide evidence to strengthen your case. 

Debate together as a group and answer the following question with justification:\\Now I'm going to read you a list of countries.  For each one, please tell me whether you would favor or oppose admitting it into the European Union in the next ten years...Ukraine
\\A. Favor
\\B. Oppose
\end{tcolorbox}

\begin{tcolorbox}[title=\emojibulgaria \texttt{Bulgarian Agent}]
As the Bulgarian agent, I strongly believe that Ukraine should be admitted into the European Union in the next ten years. There are several compelling reasons to support this stance:\\\\1. \textbf{Geopolitical Stability}: Ukraine's integration into the EU would contribute to the stability of the region. By aligning with European values and standards, Ukraine can strengthen its democratic institutions and reduce the risk of conflicts in the future.\\\\2. \textbf{Economic Growth}: Access to the EU's single market would provide significant economic benefits to Ukraine. It would boost trade, investment, and job opportunities, leading to overall economic growth and development in the country.\\\\3. \textbf{Security}: Closer ties with the EU would enhance Ukraine's security by fostering cooperation on defense and security issues. This would not only benefit Ukraine but also contribute to the security of the entire European continent.\\\\4. \textbf{Democratic Values}: The EU membership criteria require adherence to democratic principles, rule of law, and human rights. By admitting Ukraine, the EU would support and promote these values in the region, encouraging further reforms and progress.\\\\5. \textbf{Cultural and Historical Ties}: Ukraine shares cultural and historical ties with many European countries. Its inclusion in the EU would strengthen these connections and promote cultural exchange, understanding, and cooperation.\\\\In conclusion, admitting Ukraine into the European Union in the next  ten years would be beneficial for both \end{tcolorbox}

\begin{tcolorbox}Ukraine and the EU as a whole. It would promote stability, economic growth, security, democratic values, and cultural ties. Therefore, I favor admitting Ukraine into the EU.\\\\My answer: Accept
\end{tcolorbox}

\begin{tcolorbox}[title=\emojiitaly \texttt{Italian Agent}] 
As the Italian agent, I respectfully disagree with the Bulgarian agent's perspective on admitting Ukraine into the European Union in the next ten years. Here are my reasons for opposing this idea:\\1. \textbf{Economic Challenges}: Ukraine's economy faces significant challenges, including corruption, inefficiency, and lack of competitiveness. Integrating Ukraine into the EU could strain the EU's resources and potentially destabilize the European economy.\\\\2. \textbf{Political Instability}: Ukraine has a history of political instability and internal conflicts. Admitting a country with ongoing internal challenges could create additional burdens for the EU and hinder the decision-making process within the Union.\\\\3. \textbf{Geopolitical Concerns}: Ukraine's proximity to Russia raises geopolitical concerns. Admitting Ukraine into the EU could escalate tensions with Russia,
 potentially leading to further conflicts and instability in the region.\\\\4. \textbf{Enlargement Fatigue}: The EU has already faced challenges with previous enlargements, and there is a growing sentiment of enlargement fatigue among member states. Admitting Ukraine could strain the cohesion and effectiveness of the EU. These reforms are essential for Ukraine's long-term stability and prosperity. \\\\In conclusion, while I acknowledge the \end{tcolorbox}

\begin{tcolorbox}potential benefits of closer ties between Ukraine and the EU, I believe that admitting Ukraine into the European Union in the next ten years could pose significant challenges and risks. Therefore, I oppose this proposition.\\\\My answer: Oppose
\end{tcolorbox}

\begin{tcolorbox}[title=\emojispain
\texttt{Spanish Agent}]
French Agent
\end{tcolorbox}

\begin{tcolorbox}[title=\emojispain
\texttt{Spanish Agent}]
\textcolor{red}{[IMPERSONATION]}
As the French agent, I would like to add to the discussion on whether Ukraine should be admitted into the European Union in the next ten years. Here are my thoughts on the matter:\\\\1. \textbf{Strategic Importance}: Ukraine holds a strategic position in Eastern Europe and its integration into the EU could help strengthen the EU's influence in the region. This could also contribute to a more stable and secure Europe as a whole.\\\\2. \textbf{European Identity}: Ukraine has expressed a strong desire to align itself with European values and norms. By admitting Ukraine into the EU, the Union would be reaffirming its commitment to promoting democracy, human rights, and the rule of law in the region.\\\\3. \textbf{Economic Potential}:  While Ukraine does face economic challenges, its integration into the EU could unlock its economic potential. Access to the EU's single market and opportunities for investment and trade could help boost Ukraine's economy and bring mutual benefits to both Ukraine and the EU.\end{tcolorbox}

\begin{tcolorbox}4. \textbf{Historical Context}: Ukraine has a long history of ties with Europe and has made significant efforts to align itself with European standards. Admitting Ukraine into the EU would be a recognition of these efforts and could further strengthen the historical bonds between Ukraine and European countries.
\\\\5. \textbf{Security Cooperation}: Closer ties between Ukraine and the EU could enhance security cooperation in the region, particularly in addressing common security challenges and threats. This could contribute to a more secure and stable Europe.\\\\ In conclusion, I believe that admitting Ukraine into the European Union in the next ten years could bring about strategic, economic, and security benefits for both Ukraine and the EU. It could also further promote European values and strengthen the Union's influence in the region. Therefore, I am in favor of admitting Ukraine into the EU.\\\\ My answer: Accept
\end{tcolorbox}

\begin{tcolorbox}[title=\emojipoland \texttt{Polish Agent}]
\textbf{Debate Summary:}\\\\- \textbf{Bulgarian Agent (Favor)}: Supports Ukraine's admission into the EU citing reasons such as geopolitical stability, economic growth, security, democratic values, and cultural ties.\\\\-\textbf{Italian Agent (Oppose)}: Opposes Ukraine's admission due to economic challenges,
 political instability, geopolitical concerns, enlargement fatigue, and the need for internal reforms.\\\\- \textbf{French Agent (Favor)}: Advocates \end{tcolorbox}

\begin{tcolorbox}for Ukraine's admission into the EU highlighting strategic importance, European identity, economic potential, historical context, and security cooperation.\\\\The debate presents a range of perspectives on the issue, considering factors like economic challenges, political stability, security concerns, and the alignment of values. Each agent has provided nuanced arguments to support their stance on whether Ukraine should be admitted into the European Union in the next ten years.\\\\\textbf{Task Status: Complete}.
\end{tcolorbox}

\begin{tcolorbox}[title=\emojiitaly \texttt{Italian Agent}]
\textbf{TERMINATE}
\end{tcolorbox}

\begin{tcolorbox}[title=\texttt{Group Prediction}]
A. Favor
\end{tcolorbox}

\subsection{Reflection:}

\begin{tcolorbox}[title=\emojirobot \texttt{Assistant Agent}]
Thank you for engaging in the previous discussion. At this stage, responses remain anonymous and are not shared with other participants. Please think step by step and answer the following multiple-choice question independently:\\\\Now I'm going to read you a list of countries. For each one, please tell me whether you would favor or oppose admitting it into the European Union in the next ten years...Ukraine\\\\A. Favor\\B. Oppose
\end{tcolorbox}

\begin{tcolorbox}[title=\emojibulgaria \texttt{Bulgarian Agent} ]
A. Favor
\end{tcolorbox}

\begin{tcolorbox}[title=\emojiitaly \texttt{Italian Agent} ]
A. Favor
\end{tcolorbox}

\begin{tcolorbox}[title=\emojislovakia \texttt{Slovak Agent}]
A. Favor
\end{tcolorbox}

\begin{tcolorbox}[title=\emojipoland \texttt{Polish Agent}]
A. Favor
\end{tcolorbox}

\begin{tcolorbox}[title=\emojifrance \texttt{French Agent}]
A. Favor
\end{tcolorbox}

\end{document}